%% file: main.tex
\documentclass[10pt,twocolumn,letterpaper]{article}

 \usepackage{iccv}              

\usepackage{multirow}
\usepackage{multicol}
\usepackage{booktabs}
\usepackage{graphicx} 
\usepackage{caption} 
\usepackage{float}   
\usepackage[normalem]{ulem}
\useunder{\uline}{\ul}{}
\usepackage{utfsym}

\input{preamble}

\definecolor{iccvblue}{rgb}{0.21,0.49,0.74}
\usepackage[pagebackref,breaklinks,colorlinks,allcolors=iccvblue]{hyperref}

\def\paperID{1958} 
\def\confName{ICCV}
\def\confYear{2025}

\title{\model: Fantastic Video Upscalers and Where to Find Them}

\author{Zhongdao Wang$^{1}$ {}
Guodongfang Zhao$^{1}$ {}
Jingjing Ren$^{2}$ \\ 
Bailan Feng$^{1}$ {}
Shifeng Zhang$^{1}$ {}
Wenbo Li$^{1\dagger}$ 
\\
[0.1cm]
 $^1$Huawei Noah's Ark Lab \quad $^2$ HKUST (Guangzhou)
}

\begin{document}

\input{sec/0_abstract}

\input{sec/1_intro}
\input{sec/4_relwork}
\input{sec/2_method}
\input{sec/3_exp}

\input{sec/5_concl}

{
    \small
    \bibliographystyle{ieeenat_fullname}
    \bibliography{main}
}

\input{sec/X_suppl}

\end{document}

%% file: preamble.tex
\newcommand{\model}{\textsc{TurboVSR}\xspace}

\newlength\savewidth\newcommand\shline{\noalign{\global\savewidth\arrayrulewidth
  \global\arrayrulewidth 1pt}\hline\noalign{\global\arrayrulewidth\savewidth}}

\definecolor{baselinecolor}{gray}{.9}

\newcommand\blfootnote[1]{%
  \begingroup
  \renewcommand\thefootnote{}\footnote{#1}%
  \addtocounter{footnote}{-1}%
  \endgroup
}

%% file: sec/0_abstract.tex
\twocolumn[{%
\renewcommand\twocolumn[1][]{#1}%
\maketitle
\begin{center}
    \centering
    \captionsetup{type=figure}
    \vspace{-0.7cm}
    \includegraphics[width=\textwidth,height=10cm]{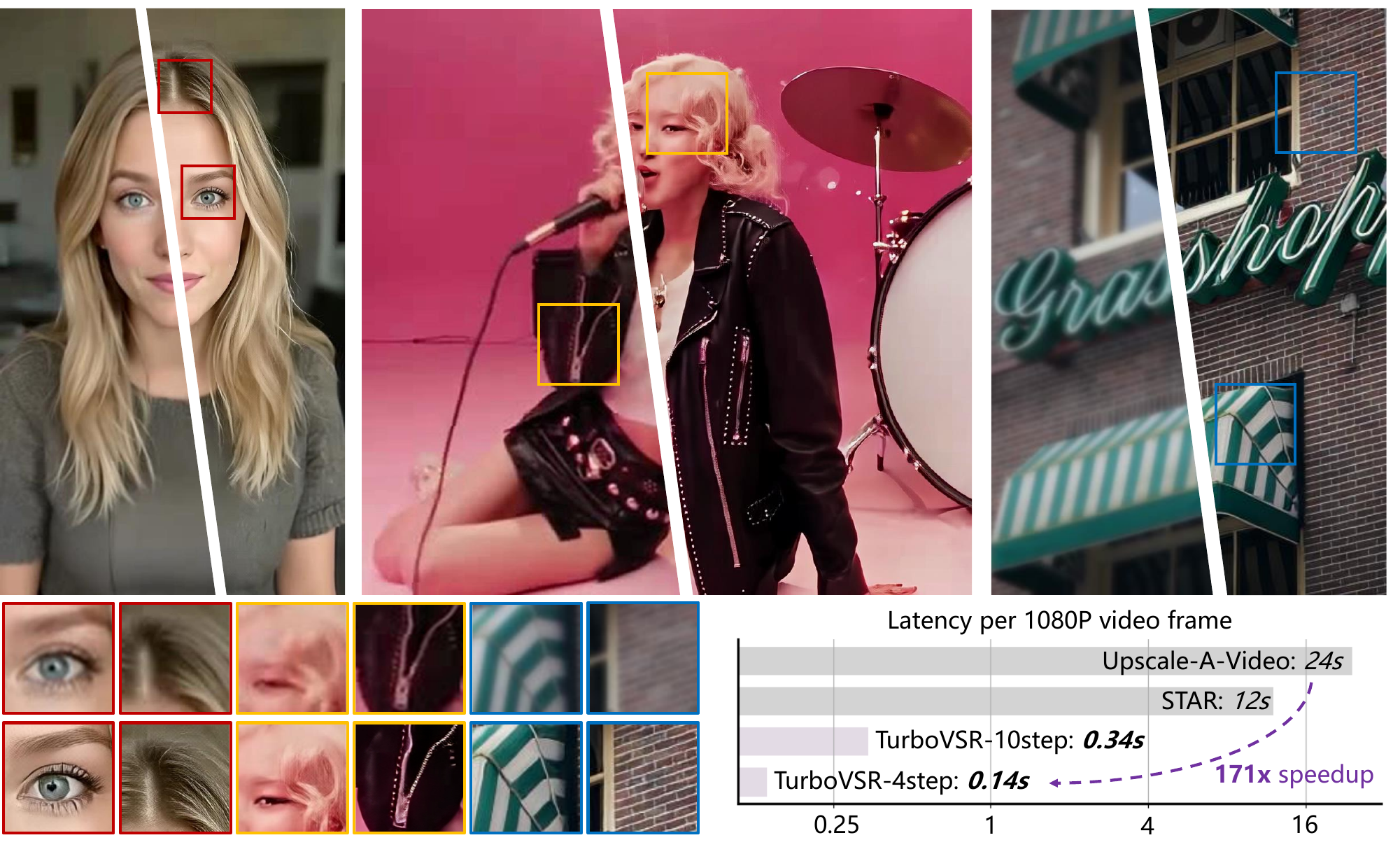}
    \vspace{-0.9cm}
    \captionof{figure}{\textbf{Qualitative results} of the proposed \model~on video super-resolution (VSR). The low-resolution (LR) inputs are shown on the left, while the super-resolved (SR) outputs are displayed on the right. For clarity, local regions are zoomed in to facilitate detailed comparison. 
While \model~demonstrates comparable performance to state-of-the-art diffusion-based VSR methods, such as Upscale-A-Video~\cite{zhou2024upscale}, it achieves a remarkable computational advantage, delivering a speedup of \textbf{over 100$\times$}. }
\end{center}%
}]

\blfootnote{$\dagger$ Corresponding author.}

\begin{abstract}
Diffusion-based generative models have demonstrated exceptional promise in the video super-resolution (VSR) task, achieving a substantial advancement in detail generation relative to prior methods.
However, these approaches face significant computational efficiency challenges.  For instance, current techniques may require tens of minutes to super-resolve a mere 2-second, 1080p video.
In this paper, we present \model, an ultra-efficient diffusion-based video super-resolution model. Our core design comprises three key aspects: 
(1) We employ an autoencoder with a high compression  ratio of 32$\times$32$\times$8 to reduce the number of tokens. 
(2) Highly compressed latents pose substantial challenges for training. We introduce factorized conditioning to mitigate the learning complexity: we first learn to super-resolve the initial frame; subsequently, we condition the super-resolution of the remaining frames on the high-resolution initial  frame and the low-resolution subsequent frames. 
(3) We convert the pre-trained diffusion model to a shortcut model to enable fewer sampling steps, further accelerating inference.
As a result, \model performs on par with state-of-the-art VSR methods, while being 100+ times faster, taking only 7 seconds to process a 2-second long 1080p video. 
\model also supports image resolution by considering image as a one-frame video. Our efficient design makes SR beyond 1080p possible, results on 4K (3648$\times$2048) image SR show surprising fine details.
\end{abstract}

%% file: sec/1_intro.tex
\section{Introduction}
\label{sec:intro}

In recent years, diffusion-based generative models have significantly advanced image and video generation tasks, also leading to transformative improvements in image and video super-resolution (ISR/VSR) tasks. These models outperform traditional methods in detail generation, greatly enhancing visual quality.
However, low computational efficiency remains a significant challenge.  For instance, models like Upscale-A-Video~\cite{zhou2024upscale} take approximately 20 minutes to process a 2-second, 1080p video on an NVIDIA H20 GPU. This inefficiency arises from two main aspects: high-resolution results in long token sequences, increasing computation; and diffusion models require multiple iterative steps, further reducing efficiency.
To address these challenges, this paper explores how to develop an efficient diffusion-based video upscaler.

We introduce \model, an ultra-efficient video upscaler  that processes 1080p resolution videos of 2 seconds in merely 7 seconds. Compared to existing methods, it achieves efficiency improvements of several tens to hundreds of times while maintaining competitive detail generation quality. Our efficient design also allows \model to handle 4K ultra-high definition resolutions (3840 × 2160) with significantly enriched content details. To realize this, we present the following core designs:

\noindent\textbf{High compression ratio autoencoder:}  
We employ LTX-VAE~\cite{hacohen2024ltx} as our autoencoder, with  a spatial compression ratio of 32$\times$ and a temporal compression ratio of 8$\times$. In comparison, the commonly used  SD-VAE~\cite{sd} offers a spatial compression ratio of 16$\times$ (8$\times$ plus 2$\times$ patchification) with no temporal compression. This results in 32$\times$ shorter token sequences, greatly reducing computational demands.
While high compression ratio autoencoders may cause performance degradation in generative tasks, in VSR, degraded video inputs serve as strong conditional information, only requiring detailed content generation. Consequently, the negative impact of high compression is relatively diminished, making it more suitable for this specific application. 
Additionally, the LTX-VAE supports both images and videos, enabling \model to perform SR for both modalities simultaneously.

\noindent\textbf{Factorized conditioning:} 
We observed that directly fine-tuning LTX-DiT (video generation model paired with LTX-VAE) for video SR yields suboptimal results, whereas its performance in  image SR is satisfactory. This discrepancy may arise from the lack of exposure to long token sequences during LTX-DiT pre-training. To optimize training efficiency, we propose a Factorized Conditioning approach to rapidly transfer the capabilities of the pre-trained model to the VSR task. We decompose VSR into two tasks: (1)~super-resolving the initial frame of an LR video, and (2)~super-resolving the entire video given the predicted initial frame and the remaining LR frames. With limited training budgets, Factorized Conditioning significantly accelerates convergence and improves final performance.

\noindent\textbf{Non-uniform shortcut models:} 
To minimize the number of sampling steps, we transformed the LTX-DiT from its original flow matching model~\cite{liuflow} to shortcut model~\cite{frans2024one}. However, this direct adaptation did not achieve the desired outcomes due to the fixed base step size and  uniform timestep sampling employed in vanilla shortcut models. In VSR, high resolution results in lengthy token sequences, necessitating a non-uniform timestep sampling strategy that shifts towards the initial timestep for optimal effectiveness~\cite{sd3}.
Consequently, we propose non-uniform sampling shortcut models that leverage multiple base step sizes along with a nearest neighbor step selection strategy to reduce errors along the ODE path. Our experiments show that this approach can reduce the sampling steps from 10 to 4 with negligible compromise to performance, achieving a 2.5$\times$ improvement in efficiency.

\noindent\textbf{Flexible inference at arbitrary  resolutions and lengths:}
We introduce a tile-wise inference strategy to enable flexible inference at arbitrary resolutions and lengths. Similar to Multi-Diffusion~\cite{multidiffusion},we split a video into multiple fixed-size tiles with small overlaps in both spatial and temporal dimension. At each diffusion step, tiles are separately processed and overlapped region are weighted blended by a Gaussian kernel. 
 Additionally, we utilized a same fixed initial noise for denoising across different tiles to further enhance consistency at the boundaries~\cite{optimalbc}. These techniques achieved strong spatiotemporal consistency, effectively eliminating visible tile boundaries.

With the support of the aforementioned efficient design, our method achieves a processing speed of 0.14 seconds per frame for 1080p video clips on a single NVIDIA H20 GPU, showing an efficiency improvement of several tens to even a hundred times compared to existing diffusion-based super-resolution methods. Meanwhile, results across multiple benchmarks demonstrate that \model maintains a competitive level of super-resolution performance compared to prior arts. 

%% file: sec/4_relwork.tex
\section{Related Work}




\noindent\textbf{Video super-resolution (VSR).}
In the field of VSR, traditional methods have focused on enhancing the quality of low-resolution videos through various temporal alignment strategies. These include independent alignment to the central frame using optical flow~\cite{xue2019video,kappeler2016video,liao2015video,ma2015handling} or adaptive kernels~\cite{tian2020tdan,li2020mucan,wang2019edvr,zhou2019spatio,li2020lapar}, and progressive alignment~\cite{sajjadi2018frame,chan2021basicvsr,chan2022basicvsr++,zhou2022revisiting} to handle long-range dependencies. Generative Adversarial Networks~\cite{goodfellow2014generative,zhang2024realviformer,chan2022investigating,xie2023mitigating,li2022best} have also been applied to VSR to generate more detailed frames. However, these methods can suffer from unstable training, over-smoothing, and temporal inconsistencies, emphasizing the need for models with strong generative priors.

\noindent\textbf{Video Generation.}
Diffusion-based models~\cite{blattmann2023align,guo2024animatediff,gupta2024photorealistic,ho2022imagenvideohighdefinition,wang2023modelscope,wang2023videocomposer,wang2024lavie,peng2024controlnext,zhang2025flashvideo} have gained prominence in text-to-video tasks. Diffusion Transformers (DiTs) replace the traditional U-Net architecture, improving the model's ability to generate high-quality videos. Building on the success of Sora~\cite{videoworldsimulators2024} and Open-Sora~\cite{zheng2024open} have been developed to produce high-fidelity video content. CogVideoX~\cite{yang2024cogvideox} introduces an expert transformer architecture with adaptive LayerNorm and progressive training techniques, producing temporally consistent videos that align well with textual descriptions. Additionally, HuanyuanVideo~\cite{,kong2024hunyuanvideo}, scaling to 13B parameters, achieves state-of-the-art performance in open-sourced video generation. To accelerate the generation process, LTX-Video~\cite{hacohen2024ltx} utilizes a high compression ratio VAE with a $32 \times 32 \times 8$ factor, though this results in compromised visual quality.

\noindent\textbf{Diffusion-based VSR.}
The promising generative capabilities of video DiTs have inspired their use in downstream tasks like VSR. Diffusion-based VSR models~\cite{li2025diffvsr,zhou2024upscale,he2024venhancer,yang2024motion,xie2025star,rota2024enhancing,chen2024learning} aim to address challenges in real-world video enhancement. For instance, the DiffVSR framework~\cite{li2025diffvsr} incorporates a multi-scale temporal attention module and a temporal-enhanced VAE decoder to capture fine-grained motion details, improving both fidelity and temporal consistency. Similarly, Upscale-A-Video~\cite{zhou2024upscale} integrates temporal layers into U-Net and VAE-decoder architectures to ensure temporal coherence, while introducing a flow-guided recurrent latent propagation module to improve video stability. Recently, STAR~\cite{xie2025star} employs a local information enhancement module and dynamic frequency loss to enrich local details. Despite these advances, the high computational cost of attention calculations and multiple sampling steps remains a significant bottleneck for inference efficiency. 

\noindent\textbf{Efficient diffusion models.}
To accelerate the execution of diffusion models, significant efforts have been dedicated to developing high compression ratio VAEs~\cite{chen2024deep,zha2024language,ma2025step}, efficient attention variants~\cite{pu2024efficient,xie2024sana,xi2025sparse}, model pruning techniques~\cite{li2023snapfusion,hu2024snapgen,yahia2024mobile}, and advanced sampling algorithms~\cite{jin2024pyramidal,frans2024one,yi2025magic} aimed at reducing the number of steps. For instance, DC-AE~\cite{chen2024deep} introduces residual auto-encoding and decoupled high-resolution adaptation to achieve a high spatial compression ratio, resulting in significant inference speedup. SANA~\cite{xie2024sana} utilizes the linearDiT approach to efficiently handle high-resolution image resolution, achieving a balance between efficiency and quality sacrifice. Shortcut models~\cite{frans2024one} condition the network on both the current noise level and the desired step size, showing promising results compared to consistency models~\cite{song2023consistency,songimproved} and reflow models~\cite{liuflow,lipmanflow}.
In this work, we focus on leveraging the powerful priors of video generation models to achieve video super-resolution, while introducing efficient optimizations in terms of token utilization and sampling efficiency.

%% file: sec/2_method.tex
\section{Method}

\subsection{Overall Architecture}
Following existing works, we define the image/video SR task as conditional generation problems. Our overall framework is inspired by ControlNeXt~\cite{peng2024controlnext}. Given a degraded image or video, we encode it into a latent code using an autoencoder, which  then injected as a condition into a diffusion model. 
The conditional latent is scaled to match the hidden state dimensions using a single linear layer. We then apply cross normalization~~\cite{peng2024controlnext}, using the mean and variance of the hidden state output from the first block of the transformer to normalize the conditional latent. Finally, we add the two components together and scale by a factor of $\sqrt{2}$ to maintain the variance of the standard distribution.
Similar to other diffusion-based SR method, an additional text condition is required to enhance generation quality. We employ Res-Captioner~\cite{rescaptioner} to generate text conditions based on the LR input. 
We adopt flow matching~\cite{liuflow} as our training framework, utilizing v-prediction as the objective during training and applying Euler solver to solve the ODE during inference.

\begin{figure*}[htb]
    \centering
    \includegraphics[width=1.0\linewidth]{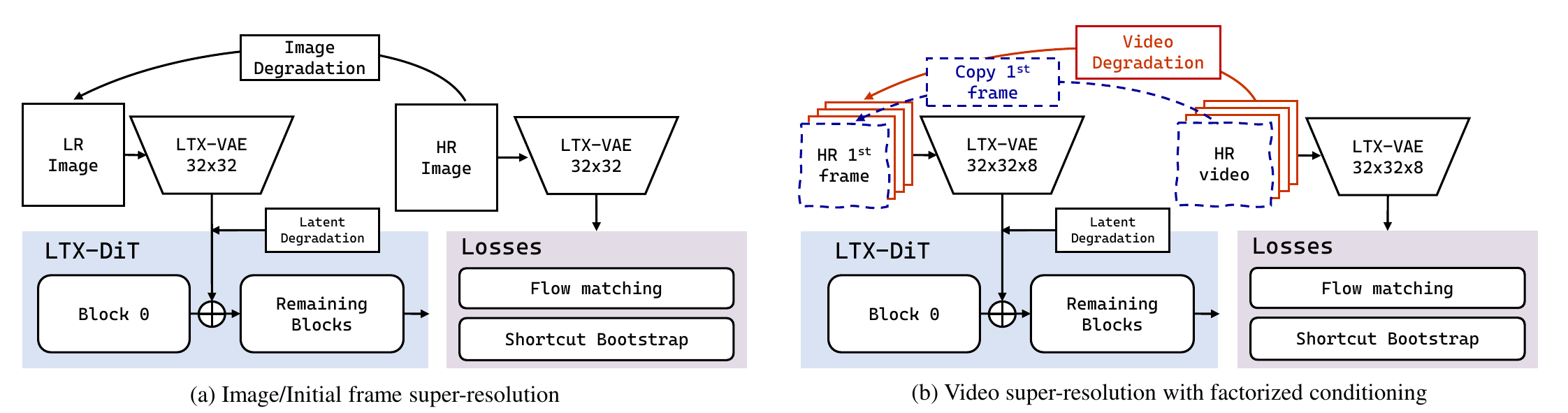}
    \caption{An illustration of the proposed factorized conditioning. We train the model to fulfill two objectives in a multi-task manner: \textbf{(a)} Image/Initial frame super-resolution; \textbf{(b)} Video super-resolution, conditioned on the super-resolved initial frame and degraded rest frames.}
    \label{fig:factorized}
\end{figure*}

\subsection{High Compression Ratio Autoencoder}

We identify the primary inefficiency of existing diffusion-based video upscalers as the excessively long sequence lengths. On one hand, the task's high resolution and long temporal characteristics necessitate long token sequences. On the other hand, most approaches rely on image autoencoders like SD-VAE~\cite{sd} or SD3-VAE~\cite{sd3}, which do not compress in the temporal dimension and exhibit limited spatial compression. Although some methods, such as STAR~\cite{xie2025star}, explore video autoencoders for VSR, their compression ratios still remain relatively conservative.

\begin{figure}
    \centering
    \includegraphics[width=\linewidth]{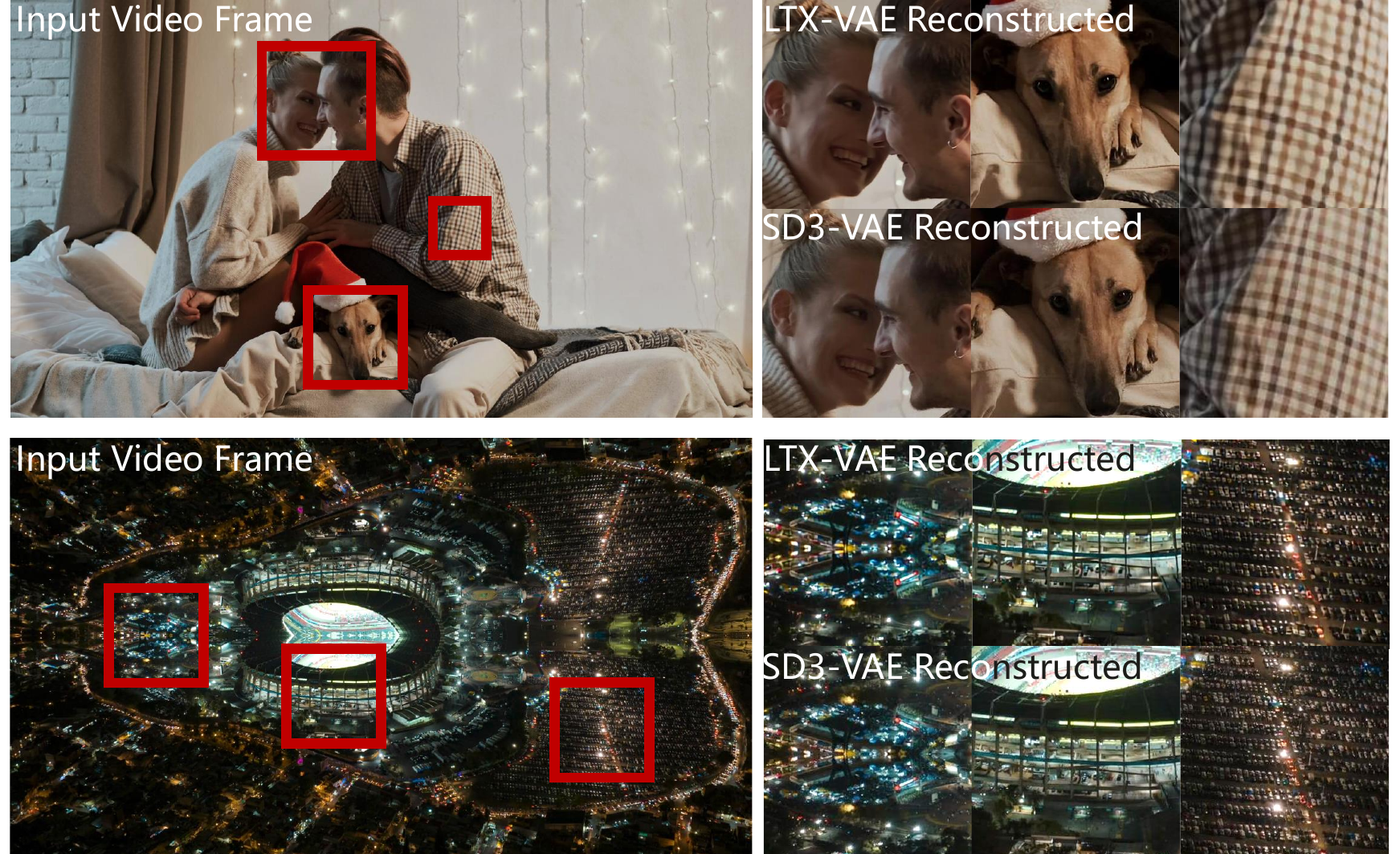}
    \vspace{-0.5cm}
    \caption{Comparison between SD3-VAE and LTX-VAE in terms of reconstruction quality. Overall, LTX-VAE is quite competitive, exhibiting only a slight deficiency in certain subtle details that are difficult to observe. }
    \label{fig:pilot_vae}
\end{figure}

We argue that a more aggressive high-compression auto-encoder is feasible for VSR. To this end, we select LTX-VAE~\cite{hacohen2024ltx}, which achieves 32$\times$ compression in the spatial dimension and 8$\times$ in the temporal dimension, with a latent dimension of 128. 
A comparison of basic parameters between SD3-VAE and LTX-VAE is listed in Table~\ref{tab:compression_ratios}.
Although the token length is reduced by 32$\times$ compared to SD-VAE, the latent dimension is correspondingly increased, theoretically possible to preserve the the detail generation capability.
To validate this, we conducted a pilot study, selecting several typical sets of 1080p videos, featuring typical scenarios such as portraits, vegetation, architecture, and fast motion. We reconstructed these videos using both LTX-VAE and SD-VAE, and present several comparative results in Figure~\ref{fig:pilot_vae}. We make the following observations.
Firstly, the differences between the reconstruction of LTX-VAE and SD3-VAE are difficult to recognize from a human intuitive perspective. Careful observation is required to discern them, \eg,  the clothing texture in Figure~\ref{fig:pilot_vae}.
Secondly, LTX-VAE demonstrates a degree of imaginative capability, resulting in a subjective perception of acceptable quality despite larger reconstruction errors.
Finally, LTX-VAE shows superiority in rendering details for portraits and animals, while performing less effectively with rapid motion and repetitive, regular textures.
In conclusion, we believe that the detail restoration capability of LTX-VAE is sufficient to support SR tasks.

\begin{table}
    \centering
\scriptsize
    \begin{tabular}{@{}lcccc@{}}
        \toprule
        &Autoencoder  & \multirow{2}{*}{Patchify} & {Latent } & Pix-to-Token  \\ 
        & Comp. Ratio & & Dim. & Comp. Ratio \\ 
        \midrule
        SD3-VAE & 8$\times$ 8$\times$1 & 2$\times$2 $\times$1& 16 & 256  \\
        LTX-VAE & 32$\times$ 32 $\times$8 & 1$\times$1$\times$1 & 128 & 8196  \\ 
        \bottomrule
    \end{tabular}
    
  
    \caption{Comparison between SD3-VAE and LTX-VAE}
    \label{tab:compression_ratios}
\end{table}

\subsection{Factorized Conditioning}

We adopt LTX-DiT~\cite{hacohen2024ltx}, the video generation model paired with LTX-VAE, as our base model, inheriting its generative capability. 
However, we observed slow convergence during training and a low success rate during inference.

We argue that the primary reason can be attributed to the mismatch between the token lengths during T2V pre-training and VSR fine-tuning. The token length for LTX-DiT during T2V training is less than 5K, while our fine-tuning for VSR requires 9K, corresponding to a 1024px, 49-frame video clip. This suggests an out-of-distribution (OOD) transfer is needed for the pre-trained model, inhibiting effective fitting within a short timeframe. To validate this, we trained an image SR model (1024px, 1 frame)  based on LTX-DiT with the token length reduced to 1024. The results demonstrated a significant improvement in both SR success rate and the quality of detail generation.

To enable efficient training and leverage the prior knowledge of the base model, we propose Factorized Conditioning, illustrated in Figure~\ref{fig:factorized}. We employ a multi-task approach to alternately train a shared LTX-DiT model, optimizing two objectives: first, single-image SR, for upscaling  the initial frame of the video; second, video SR, which conditions on the latent representation of the predicted initial frame and the latents of the remaining degraded frames, for upscaling the entire video.
The image SR component efficiently inherits the generative capabilities of the base model with minimal training, while the video SR component benefits from strong signals provided by the image SR results. The latter only requires learning frame-to-frame correspondence~\cite{wang2023unsupervised,wang2019learning,jabri2020space} to propagate details across frames, significantly reducing the learning difficulty, thus accelerating convergence and improving overall performance. We present an ablation study of factorized conditioning in Section~\ref{sec:abl}.

\subsection{Non-uniform Shortcut Model}
 With the support of factorized conditioning, the high compression autoencoder reduces computational costs by tens of times. To further enhance efficiency, we identify opportunities in reducing the number of sampling steps.
We adapt LTX-DiT to a shortcut model~\cite{frans2024one} to enable fewer sampling steps. Compared with distillation-based methods~\cite{dmd,song2023consistency}, shortcut model is simpler, avoiding multiple teacher-student training phases, and more flexible,  supporting many-step generation rather than a fixed step number.

\paragraph{Shortcut models.} 
Flow matching model formulates the generation problem as learning an ordinary differential equations (ODEs) that maps from the noise distribution to the data distribution. Defining $x_t$  as the linear interpolation between the data $x_0 \sim \mathcal{D}$ and the noise $x_1 \sim \mathcal{N}(0, I)$, the velocity $v_t$ points from the noise towards the data: 
\begin{equation}
    x_t = (1-t)  x_0 + t x_1 \quad\text{and} \quad v_t = x_1 - x_0.
\end{equation}
The flow matching objective is to regress the average direction of $v_t$ by a neural network parameterized with $\theta$ via objective 
\begin{equation}
    \mathcal{L}_{\texttt{flow}} = \mathbb{E}_{x_0\sim\mathcal{D}, x_1 \sim \mathcal{N}(0, I)}[\Vert v_\theta(x_t, t) - (x_1 - x_0)\Vert^2].
\end{equation}
Since the leaned $v_\theta$ is the average velocity, even perfectly trained ODE fall shorts in predicting accurate paths under few-step setting. Shortcut models~\cite{frans2024one} are proposed to resolve the few-step ambiguity by introducing the step size $d$ as an additional condition term by learning a \emph{shortcut}
\begin{equation}
    x'_{t+d} = x_t + v_\theta (x_t, t, d)d.
\end{equation}
The shortcut can be learned by enforcing a self-consistency property: one shortcut step equals two steps of half the size:
\begin{equation}
    v_\theta (x_t, t, 2d) =\frac{1}{2} ( v_\theta(x_t, t, d) + v_\theta (x'_{t+d}, t, d))
\end{equation}
And the corresponding loss function is:
\begin{equation}
\footnotesize
        \mathcal{L}_{\texttt{sc}} = 
        \mathbb{E}_{x_0\sim\mathcal{D}, x_1 \sim \mathcal{N}(0, I), (t,d) \sim p(t,d)}  [\Vert v_\theta(x_t, t, 2d) - \text{sg}(v^{*})\Vert^2], 
\end{equation}
where $v^{*} =\frac{1}{2} ( v_\theta(x_t, t, d) + v_\theta (x'_{t+d}, t, d)) $ is the target, $\text{sg}(\cdot)$ indicates stop gradient, and $p(t,d)$ is the distribution of timestep $t$ and step size $d$ to sample from. 
During training, we combine $\mathcal{L}_{\texttt{flow}}$ and $\mathcal{L}_{\texttt{sc}}$ by alternative training across mini-batches.

\paragraph{Uniform bootstrap paths.}
The vanilla shortcut models employ a uniform sampling of timestep $t$ from ${\{\frac{m}{M}\}_{m=0}^{M-1}}$, where  $M$ denotes the total sampling steps which can be selected from $\{2^{0}, 2^1, ..., 2^7\}$.
Correspondingly, the step size is randomly sampled from $d \in \{2^{0}, 2^{-1}, ..., 2^{-7}\}$. 
This sampling strategy essentially quantizes the step size to a minimum unit of $2^{-7}$ . Consequently, it employs a bootstrap approach to learn larger step size shortcuts through a cumulative aggregation of smaller step size shortcuts.

\paragraph{Non-uniform bootstrap paths.}
We find that uniform bootstrap sampling for \( t \) and \( d \) performs poorly in video SR. A significant reason is that vanilla shortcut models were only tested on images with a resolution of 256px, whereas we need to handle high resolution and long sequences. SD3~\cite{sd3} indicates that under high-resolution conditions,  \( t \) must sampling shift towards \( t=1 \) (higher noise) to maintain a comparable signal-to-noise ratio, thereby making the initial distribution closer to a standard Gaussian.
Consequently, we adhered to this principle during both training and testing, employing non-uniform $t$ sampling. This led to a greater variety of possible values for \( d \), exceeding its fixed range during training. We found that selecting the closest value within the set \( \{2^{0}, 2^1, \ldots, 2^7\} \) effectively alleviates this issue.

However, this approach incurs a notable error for larger step sizes, such as \( d \in (0.5, 1) \). Thus, we incorporated additional \( d \) sampling, defining a set \( T\in \{0.6, 0.7, 0.8, 0.9, 1\} \). For each \( T \), we introduced \( d \in \{2^{0} T, 2^{-1} T, \ldots, 2^{-7} T\} \). The final \( d \) is selected by randomly choosing a \( T \) and then \( d \) from the corresponding range.
The non-uniform bootstrap paths effectively adapt the shortcut model to high resolution. Section~\ref{sec:abl} presents ablation results.

\begin{figure}
    \centering
    \includegraphics[width=\linewidth]{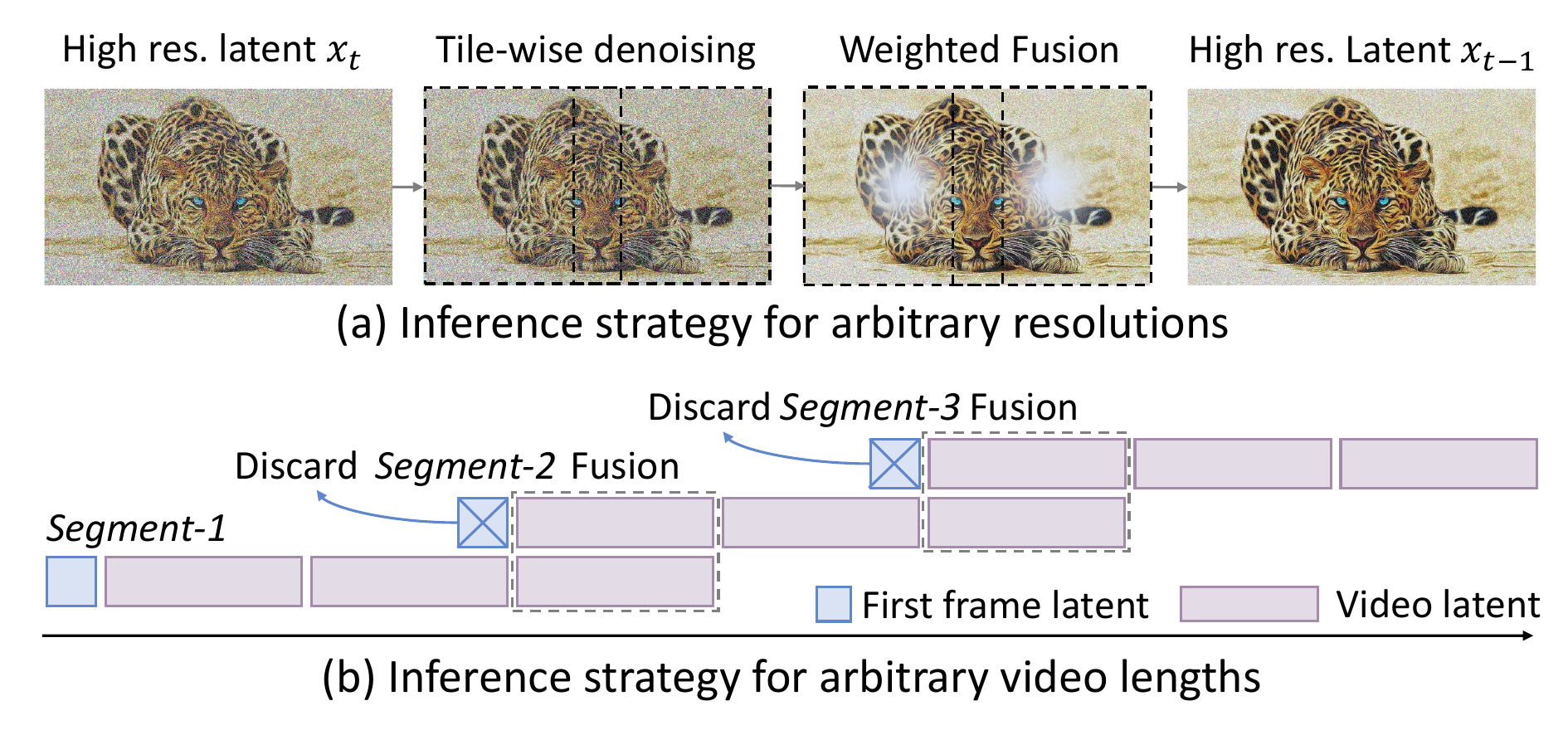}
    \caption{We split videos into fixed-length segments with overlaps. \textbf{(a)} Inside a segment we apply multi-diffusion~\cite{multidiffusion} to enable arbitrary resolutions.
    \textbf{(b)} Segments are fused to enable arbitrary length. We discard first-frame latent in the fusion.
    }
    \label{fig:aral}
\end{figure}

\subsection{Inference at Arbitrary Resolutions and Lengths}

\noindent\textbf{Arbitrary Resolution.} 
During model training, we set the input and output sizes to a fixed resolution and length. For inference, we employ a strategy akin to Multi-Diffusion~\cite{multidiffusion} to facilitate arbitrary resolution inference. Specifically, the input video is first partitioned into multiple segments with a length equal to the model input.
Then for each segment, it is 
partitioned into multiple tiles. The resolution of each tile is consistent with the fixed resolution  of the model.
Tiles are separately denoised. After each denoising step, we aggregate the tile-wise latent representations through weighted averaging to construct the overall latent of the segment. The fusion weights are given by a Gaussian kernel, which helps mitigate boundary discontinuities. In the subsequent denoising step, we repeat the procedures of tile partitioning, denoising, and fusion. To enhance consistency across different tiles, we employ fixed random noise throughout the inference process. Prior research~\cite{optimalbc} indicates that an optimal, fixed initial noise exists for diffusion ODEs. In our approach, we empirically select a noise that exhibits optimal performance.

\noindent\textbf{Arbitrary Length.}
We facilitate arbitrary-length inference through a segment fusion approach, as illustrated in Figure~\ref{fig:aral}. Our method, based on a video autoencoder, introduces specific challenges during fusion. Unlike image autoencoders, video autoencoders process the initial frame distinctly: it corresponds to a single latent representation, while subsequent frames aggregate eight into one.
This distinction complicates latent fusion, precluding the simple averaging of overlapping latents. To address this, we ensure a minimum overlap of 9 frames (1 + 8) between clips and discard the latent of the first clip frame during fusion, merging only the video latents. This modification achieves satisfactory fusion results. Furthermore, decoding the latent representation of the extended video must also be conducted in segments to maintain a consistent memory footprint.

%% file: sec/3_exp.tex
\section{Experiments}
\subsection{Training Details}
\noindent\textbf{Training data.} 
We use 5M in-house high-resolution images to train image SR and 0.5M videos for video SR, respectively. 
Both image and video resolutions exceed 1080p, with video lengths limited to 5 seconds.
 We pre-process videos to $1024 \times1024$ resolution, 49 frame clips for 1080p VSR.
We follow existing works~\cite{zhou2024upscale,xie2025star,yang2024motion} and employ the RealESRGAN~\cite{realesrgan} degradation to generate low-resolution conditions. 
To mitigate the scarcity of high-quality video data, we augmented the video dataset using image data. 
Details on data augmentation and other training strategies are deferred to the supplementary material.



\begin{table}[]
\scriptsize
    \centering
    \begin{tabular}{l | c c c c c }

         Method & Success(\%)& Detail& Fidelity& Sharpness& Stability\\
         \shline
         RealVIFormer  & 0.9& 22.83& 57.35& 45.58&  70.16\\
         MGLDVSR  & 0.825& 32.60& 55.88& 54.41&  44.37\\
         STAR & 0.95& \textbf{70.65}& \textbf{67.64}&  70.59&  79.03\\
         Upscale-A-Video  & 0.9& 55.43& 57.35& 54.41& 59.67\\
         \textbf{\model }  & \textbf{0.975}& {\ul 64.13}& {\ul 61.76}& \textbf{75.0}& \textbf{83.87}\\

    \end{tabular}
    \caption{\textbf{User study} suggests the proposed \model exhibits best success rate, good detail generation capability, fidelity, and best sharpness and stability among compared methods.}
    \label{tab:user_study}
\end{table}

\begin{table*}[ht]
    \centering
    \footnotesize

    \begin{tabular}{@{}ll|cc|ccccccc@{}}
        \multirow{2}{*}{Dataset}  & \multirow{2}{*}{Metric}   & RealBasic-& RealVI- & Upscale-A- & MGLD &  VEnhancer & STAR & \model & \model \\
         &    & VSR~\cite{chan2022investigating} & Former~\cite{zhang2024realviformer} & Video~\cite{zhou2024upscale} & VSR~\cite{yang2024motion} &  \cite{he2024venhancer} & \cite{xie2025star} & 10--step & 4-step \\
        
        \shline
        \multirow{5}{*}{SPMCS~\cite{spmcs}} 
        & DOVER$\uparrow$  & 66.17  &  71.19 & 75.82 & 67.75 & 65.76 & 48.75 & \textbf{79.29} & \underline{79.15}\\
         & MUSIQ$\uparrow$ &  64.49 & 67.01 & \textbf{68.28} & 63.68 & 59.94 & 48.06 & \underline{67.33} & 67.05\\
         & NIQE$\downarrow$ & 4.751  & 4.750 & \textbf{3.814} & 4.204 & 4.818 & 6.114 & 4.120 & \underline{4.113}\\
         & PSNR$\uparrow$ & 23.77 & 27.08 & \underline{24.12} & \textbf{26.28} & 19.04 & 23.92 & 22.10 & 22.15\\
        & SSIM$\uparrow$ & 0.690 & 0.799 & 0.660 & \textbf{0.782} & 0.472 & \underline{0.689} & 0.687 & 0.683\\
        \hline
        \multirow{5}{*}{UDM10~\cite{udm10}} 
        & DOVER$\uparrow$ & 75.01 & 82.06 &\textbf{81.74}  & 75.36 & 73.86 & 67.32 &  \underline{80.59} &  79.81\\
         & MUSIQ$\uparrow$ & 61.33 & 60.05 & \textbf{62.53} & 54.95 & 54.05 & 47.31 &  \underline{59.21} &  56.07\\
         & NIQE$\downarrow$ & 4.989 & 4.728  & \underline{4.166} & 4.442 & 5.154 & 6.063 & \textbf{3.991} &  4.532\\
         & PSNR$\uparrow$ & 24.03 & 30.88 & 27.49 & \textbf{30.04} & 22.18 & \underline{28.44} &  24.23 &  22.71\\
        & SSIM$\uparrow$ & 0.751 & 0.905 & 0.792 & \textbf{0.871} & 0.697 & \underline{0.850} &  0.769 &  0.757\\
        \hline
        \multirow{3}{*}{VideoLQ~\cite{chan2022investigating}} 
        & DOVER$\uparrow$ & 70.78 & 70.93 & 69.35 & \textbf{74.16} & \underline{ 72.63} & 68.30 &  69.75 &  70.44\\
         & MUSIQ$\uparrow$ & 68.93 & 67.12 & 65.61 & \textbf{68.96} & 67.01 & 63.65 &  \underline{67.06} &  66.58\\
         & NIQE$\downarrow$ & 5.47 & 4.902 & \underline{4.433} & 5.18 & \textbf{4.42} & 5.091 &  5.328 & 4.99\\
        \hline
         \multicolumn{2}{l|}{\hspace{-0.25cm } \textbf{Latency (ms/1080p frame)} $\downarrow$} 
         & 141 & 91  & 24,032 & \emph{37,822}  & 10516 & 12,096 &  \underline{337} & \textbf{140}  \\
    \end{tabular}
    \caption{Quantatitive comparisons on VSR benchmarks. \model 4-steps runs at 140 ms per 1080p frame, around 100+ times faster than other diffusion based VSR models, while the performance maintains quite competitive. \model performs best on SPMCS dataset, and is comparable with state-of-the-arts on UDM10 and VideoLQ. Best results of diffusion-based models are \textbf{bolded}, and second best results are \underline{underlined}.}
    \label{tab:vsr}
\end{table*}

\subsection{User Study}
We first compared the super-resolution performance of \model with existing methods through user study. A set of low-quality real network videos was randomly selected, including diverse content such as portraits, flora and fauna, and architecture. These videos were processed using different super-resolution techniques and then distributed to users for scoring. The evaluation criteria included success rate, detail generation capability, fidelity, sharpness, and video stability. The success rate required a binary score (0/1) for each video, with a score of 1 indicating significant super-resolution improvement and 0 otherwise, resulting in an average score across all videos. The other dimensions were rated on a scale from 1 to 5. To mitigate user bias, we normalized the scores to range [0,100] for each user. The final results are presented in Table~\ref{tab:user_study}.

The results indicate that \model exhibits superior generalization compared to existing diffusion-based methods. We attribute this strong generalization to the information degradation effect of the high-compression autoencoder, which makes the model less prone to overfitting to training degradation. In terms of detail generation, most diffusion-based methods outperform the non-diffusion method RealVIFormer, with our \model achieving second best detail generation.  Additionally, \model demonstrates excellent sharpness and stability.

\subsection{Results on VSR benchmarks}

\noindent\textbf{Experiment setting:} 
We conducted quantitative comparisons on three commonly used VSR benchmark datasets: SPMCS~\cite{spmcs}, UDM10~\cite{udm10}, and VideoLQ~\cite{chan2022investigating}. Following prior works, we generated LR videos for SPMCS and UDM10 using RealESRGAN degradation, while for VideoLQ, we utilized the provided LR videos. The evaluation metrics included non-reference metrics DOVER~\cite{dover}, MUSIQ~\cite{musiq}, and NIQE~\cite{NIQE}, as well as reference metrics PSNR and SSIM. Given that video super-resolution is an ill-posed problem without a standard ground truth, our focus primarily centered on the non-reference metrics. Additionally, we compared the running speed, measuring the latency per frame at 1080p resolution on an NVIDIA H20 GPU.

\noindent\textbf{Results.}
Table~\ref{tab:vsr} presents the results. Several observations can be made:
First, non-diffusion-based methods (RealBasicVSR, RealVIFormer) exhibit low latency, requiring approximately 100 milliseconds per frame, and demonstrate good fidelity with high reference metric scores; however, their detail generation capability is relatively weak (as indicated by DOVER/NIQE scores on SPMCS).
In contrast, diffusion-based methods (all others except the first two) are significantly slower, requiring over 10 seconds per frame, and generally show lower fidelity with poor reference metric scores. Nevertheless, they excel in detail generation compared to non-diffusion-based methods, as evidenced by lower NIQE scores.
Finally, as a diffusion-based method, our \model achieves comparable or even superior generation capabilities (see NIQE, MUSIQ, DOVER) while maintaining a  speed even competitive to non diffusion-based methods, with a 4-step inference time of only 140 milliseconds.
Overall, \model strikes an optimal balance between generation capability and computational cost in the VSR task, significantly enhancing the practical value of diffusion-based methods.

\subsection{Results on ISR benchmark}

\begin{table}[h]
\scriptsize
    \centering
    \begin{tabular}{l | ccccc}
                  & PSNR & SSIM & LPIPS & NIQE & MUSIQ \\
        \shline
         ResShift~\cite{resshift} &  23.94 & 0.657 & 0.298 & 7.371 & 52.96 \\
         StableSR~\cite{stablesr} & 24.69 & 0.720 & 0.200 & 6.282 & 54.74 \\
         \model (10 step) & 24.06 & 0.686 & 0.220 & 5.874 & 54.69 \\
          \model (4 step) & 24.51 & 0.704 & 0.212 & 6.042& 53.01 \\
    \end{tabular}
    \vspace{-0.3cm}
    \caption{Comparison with image SR methods.}
    \label{tab:isr}
    \vspace{-0.2cm}
\end{table}
Our architecture supports both video and image super-resolution, so we also evaluated image super-resolution metrics on the DRealSR~\cite{drealsr} benchmark, comparing \model with two existing diffusion-based methods, Reshift and StableSR. Similarly, we used RealESRGAN degradation to generate LR samples. The results are presented in Table~\ref{tab:isr}. We observed that \model achieves performance comparable to existing image super-resolution methods, both in reference and non-reference metrics, indicating that high compression ratio autoencoders have significant potential in image SR task as well.

\subsection{Higher Resolution Beyond 1080p}
\begin{figure}
    \centering
    \includegraphics[width=\linewidth]{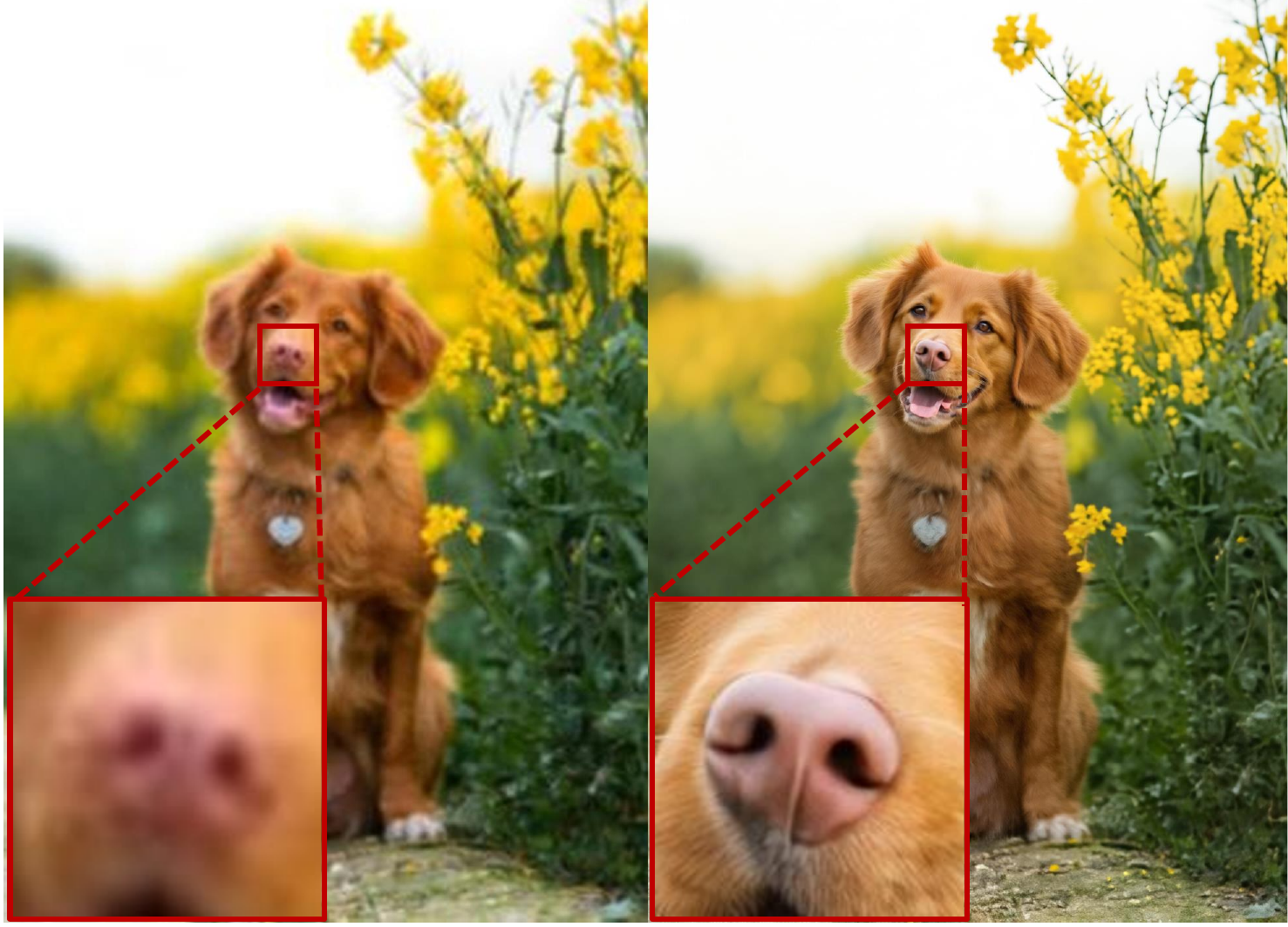}
    \caption{Example on 4K resolution image SR.}
    \label{fig:4k_demo1}
\end{figure}

Extensive efficiency optimization enables \model to achieve super-resolution beyond 1080p easily. We trained a 4K resolution image super-resolution version of \model to show this capability.
The LTX-DiT base model we used has not been trained at such high resolutions; therefore, we incorporated a 4K text-to-image pretraining stage to enhance the base model’s generative capability at this resolution.
It is worth noting that 4K video super-resolution is also feasible; however, due to computational budget constraints and a lack of high-definition video data, we leave this for future work. An example is presented in Figure~\ref{fig:4k_demo1}, with more examples available in the supplementary materials. The short side length of the input LR image is 256, while the short side length of the SR image output is 2048, yielding an  upscaling factor of 8.
Observations indicate that \model effectively leverages the generative capabilities of the pretrained model, realistically adding details of the dog's fur and facial features, while maintaining a high level of fidelity overall. We noted that the 4K image version of \model performs particularly well in animal and portrait scenes, likely due to the greater volume of data for these scenarios in the pretraining phase. 

\begin{figure*}[t]
    \centering
    \includegraphics[width=\linewidth]{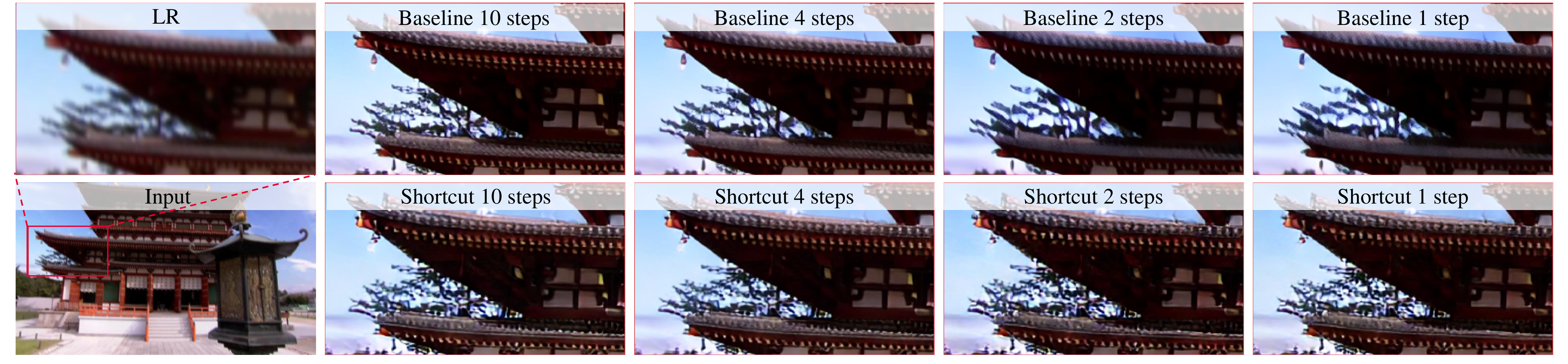}
    \vspace{-0.4cm}
    \caption{Qualitative comparison between using and not using the shortcut bootstrap loss.}
    \label{fig:sc}
    
\end{figure*}

\subsection{Ablation Study}
\label{sec:abl}
\noindent\textbf{Video condition \vs Factorized condition.} 
To validate the effectiveness of the factorized condition, we conducted a comparative experiment. For the baseline model, we utilized standard, non-factorized videos as the condition. All other training and testing conditions remained unchanged, and the inference steps are set to 10. We compared the factorized condition version of the model against the baseline, evaluating quantitative metrics on the SPMCS dataset. The results are presented in Table~\ref{tab:factorized}.
It is evident that the factorized condition significantly outperforms the standard video condition baseline in terms of NIQE, MUSIQ and DOVER, suggesting superior generation capability. Video condition baseline shows better reference metrics like PSNR and SSIM. This is mainly attributed to more failure of super resolution. and the  blurry outputs leads trivial improvements in reference metrics.
This comparison indicates that this first-frame factorization approach effectively accelerates convergence, making it easier to achieve superior performance with the same training cost.

\begin{table}[]
\footnotesize
    \centering
    \begin{tabular}{l | c c c c c }
         Condition & PSNR$\uparrow$ &  SSIM$\uparrow$  & NIQE$\downarrow$ & MUSIQ$\uparrow$  & DOVER$\uparrow$  \\
         \shline
         Video &  \textbf{24.81} & \textbf{0.759} & 6.162 & 44.87 & 34.23 \\
         Factorized & 22.15 & 0.683 & \textbf{4.113} & \textbf{67.05} & \textbf{79.15} \\
    \end{tabular}
    \caption{Ablation on factorized conditioning.}
    \label{tab:factorized}
    \vspace{-0.3cm}
\end{table}

\noindent\textbf{With \vs without shortcut.}
Subsequently, we verified the superior generation capability of the shortcut model. We evaluate on the SPMCS dataset, and comparing NIQE, MUSQ and DOVER scores. For comparison, we use a baseline model trained with a single flow matching loss. Results are shown in Figure~\ref{fig:sc}.
Overall, models trained with the shortcut bootstrap loss consistently outperform the baseline across various inference steps. Notably, the results obtained with 4 inference steps using the shortcut model closely match the quality of baseline model  achieved with 10 inference steps. Moreover, even with very few steps, such as 1 or 2 steps, the generative performance remains commendable.   Qualitative comparisons in Figure~\ref{fig:sc} also suggest similar results .


\begin{table}[]
\footnotesize
    \centering
    \resizebox{\linewidth}{!}{
    \begin{tabular}{l|l|ccc}
         & Method   & NIQE$\downarrow$& MUSIQ$\uparrow$& DOVER$\uparrow$\\ \shline
10 steps& Baseline      & 4.28& 66.35& 78.08\\
      & Shortcut & {\ul 4.12}& \textbf{67.33}& \textbf{79.29}\\ \hline
4 steps& Baseline      & 4.84& 62.14& 74.09\\
      & Shortcut &  \textbf{4.11}& {\ul 66.33}& {\ul 79.15}\\ \hline
2 steps& Baseline      & 5.82& 58.45& 69.55\\
      & Shortcut & 4.33& 66.13& 74.59\\\hline
 1 step& Baseline      
& 5.59& 58.04& 68.31\\
 & Shortcut & 4.42& 66.95& 77.08\\\end{tabular}
}    \caption{Quantitative ablation on the  shortcut loss.}
    \label{tab: sc}
\end{table}
        

\noindent\textbf{Non-uniform \vs uniform shortcut bootstrap}
We then evaluate the effectiveness of non-uniform shortcut bootstrap. We compare the timestep \( t \) and the step size \( d \) under both non-uniform and uniform sampling settings. For all evaluation, we set the sampling steps to 4. To assess video quality, we employed three perceptual evaluation metrics: perceptual similarity (LPIPS\cite{zhang2018unreasonable}) at the frame level, temporal consistency ($\mathrm{\textit{E}}_{warp}^{*}$\cite{lai2018learning}), and video clarity (DOVER) at the video level. Results in Table~\ref{tab:non-uniform} indicate that the non-uniform sampling strategy for the timestep 
 \( t \) significantly enhances video quality, while the non-uniform sampling of the step size \( d \) contributes to the stability in the generated videos. The optimal results were achieved when both non-uniform strategies are applied in conjunction. 
 

\begin{table}[]
\footnotesize
    \centering
    \resizebox{\linewidth}{!}{
   \begin{tabular}{cccc|ccc}
\begin{tabular}[c]{@{}c@{}}Uniform\\ \( t \)\end{tabular} & \begin{tabular}[c]{@{}c@{}}Non-uniform\\ \( t \)\end{tabular} & \begin{tabular}[c]{@{}c@{}}Uniform\\ \( d \)\end{tabular} & \begin{tabular}[c]{@{}c@{}}Non-uniform\\ \( d \)\end{tabular} & LPIPS$\downarrow$ &  $\mathrm{\textit{E}}_{warp}^{*}$ $\downarrow$ & DOVER$\uparrow$ \\ \hline
\usym{1F5F8}                                                   &                                                         & \usym{1F5F8}                                                   &                                                         & 0.207 & 3.77  & 71.44 \\
\usym{1F5F8}                                                   &                                                         & \textbf{}                                           & \usym{1F5F8}                                                       & 0.205 & 3.71  & 70.81 \\
                                                    & \usym{1F5F8}                                                       & \usym{1F5F8}                                                   &                                                         & 0.218 & 4.42  & 78.14 \\
                                                    & \usym{1F5F8}                                                       &                                      & \usym{1F5F8}                                                       & \textbf{0.184}& \textbf{3.57}& \textbf{79.15}\end{tabular}
  }  
  \caption{Ablation on uniform and non-uniform shortcut sampling.} 
  \label{tab:non-uniform}
  \vspace{-0.3cm}
\end{table}

%% file: sec/5_concl.tex
\section{Conclusion}
In this paper, we propose \model, an diffusion based video super-resolution model with ultra efficient designs. 
We provide three key technical contributions: high compression ratio autoencoder for reducing tokens; factorized conditioning for easy convergence; and non-uniform shortcut model for fewer-step sampling. 
Compared with existing diffusion-based VSR models, we reduce the computation cost by 100+ times, while results on benchmarks are still comparable. We hope our design, especially the usage of high compression ration autoencoder,  to inspire future research in this direction.

%% file: sec/X_suppl.tex
\clearpage
\setcounter{page}{1}
\maketitlesupplementary

\section{Training details}
\noindent\textbf{Video data augmentation.}
Due to the scarcity of high-quality, high-resolution videos and the availability of high-quality, high-resolution images, we design a video data augmentation method based on static images.  
This is achieved by generating pseudo-videos from static images using affine transformations, including random translation, rotation, and scaling.  
We carefully tune the ranges of these transformation parameters to ensure the pseudo-videos exhibit motion comparable to real videos .

\noindent\textbf{Training Strategy.} 
Image-video mixed training is achieved through alternating batches. To mitigate overfitting to training degradation, we employed  following improvements: (1) add 0-300 step random DDPM noise degradation to the condition latent; (2) fix the FFN of the network and trained other parameters; (3) while training image/video super-resolution (ISR/VSR), we also train text-to-image and text-to-video generation by randomly dropping the LR condition, for preserving the model's generation capabilities.

\vspace{-0.2cm}
\section{Qualitative Comparison}
\begin{figure*}
    \centering
    \includegraphics[width=\linewidth]{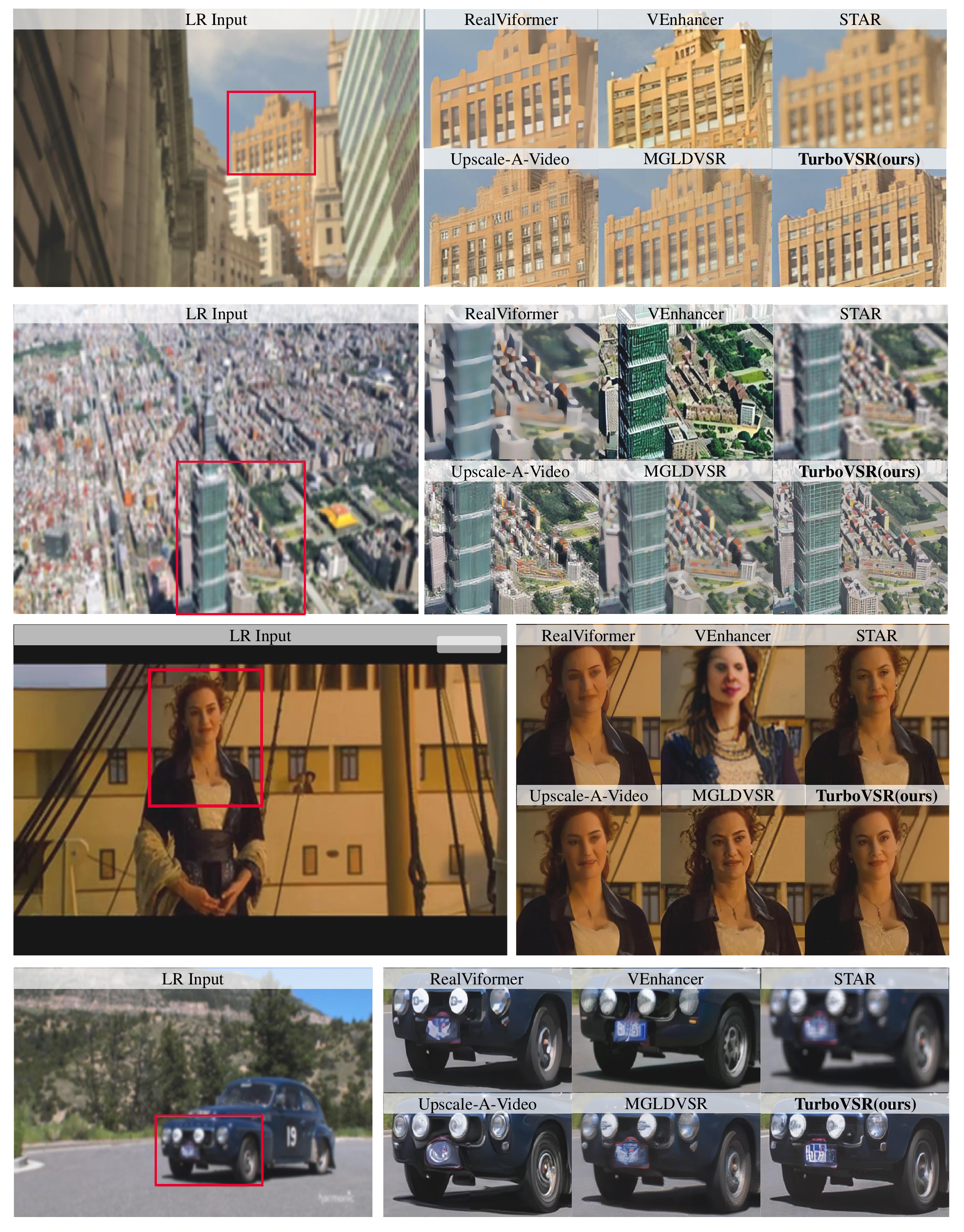}
    \caption{Qualitative comparison with existing VSR methods.}
    \label{fig:comp}
\end{figure*}

We show several qualitative comparison with existing VSR methods in Figure~\ref{fig:comp}.
Overall, \model presents detail generation capability on par with or even superior to state-of-the-art methods.
\vspace{-0.2cm}

\section{Details on 4K Resolution Image SR}

For 4K image SR, we divid the training  into two  stages, both of which utilize the same training dataset: a private 4K image dataset containing approximately 2 million samples. This dataset includes diverse contents such as portraits, landscapes, and animals, most of which are high-quality professionally generated content (PGC).  

\begin{figure}
    \centering
    \includegraphics[width=\linewidth]{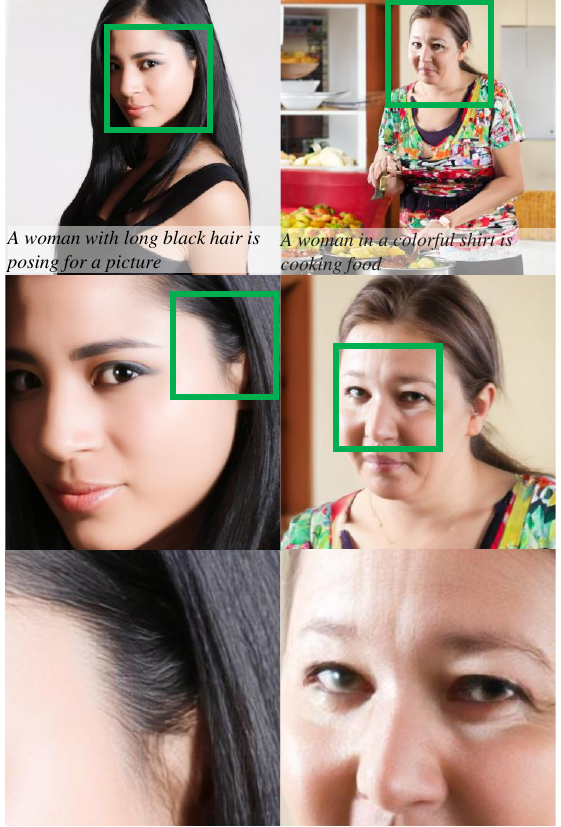}
    \caption{Examples of our high-resolution T2I generation pre-training (2048$\times$2048). Based on our high compression ratio autoencoder, we show satisfactory detail generation capability can be achieved.}
    \label{fig:4k_t2i}
    \vspace{-0.4cm}
\end{figure}

In the first stage, we train a our model on text-to-image (T2I) generation task at a resolution of 2048$\times$2048. The primary reason for this choice is that the LTX pre-trained model has not been trained at such high resolutions, making it unsuitable as a direct pre-training model for 4K super-resolution. We initialized the training with the official weights of LTX-DiT and fine-tuned it for 35K iterations with a batch size of 256. Figure~\ref{fig:4k_t2i} shows two examples generated during this stage. We observed that the model trained in this phase is capable of generating high-quality details, especially in domains such as portraits and animals, with sharp and rich details in hair and facial features. However, similar to other text-to-image generation models, it struggles with generating anatomically correct limbs and fingers, and these issues are slightly more pronounced compared to state-of-the-art models. We attribute these limitations to the lack of high-resolution training in LTX and the inherent challenges of learning from highly compressed, high-dimensional latent spaces.  

\begin{figure*}
    \centering
    \includegraphics[width=\linewidth]{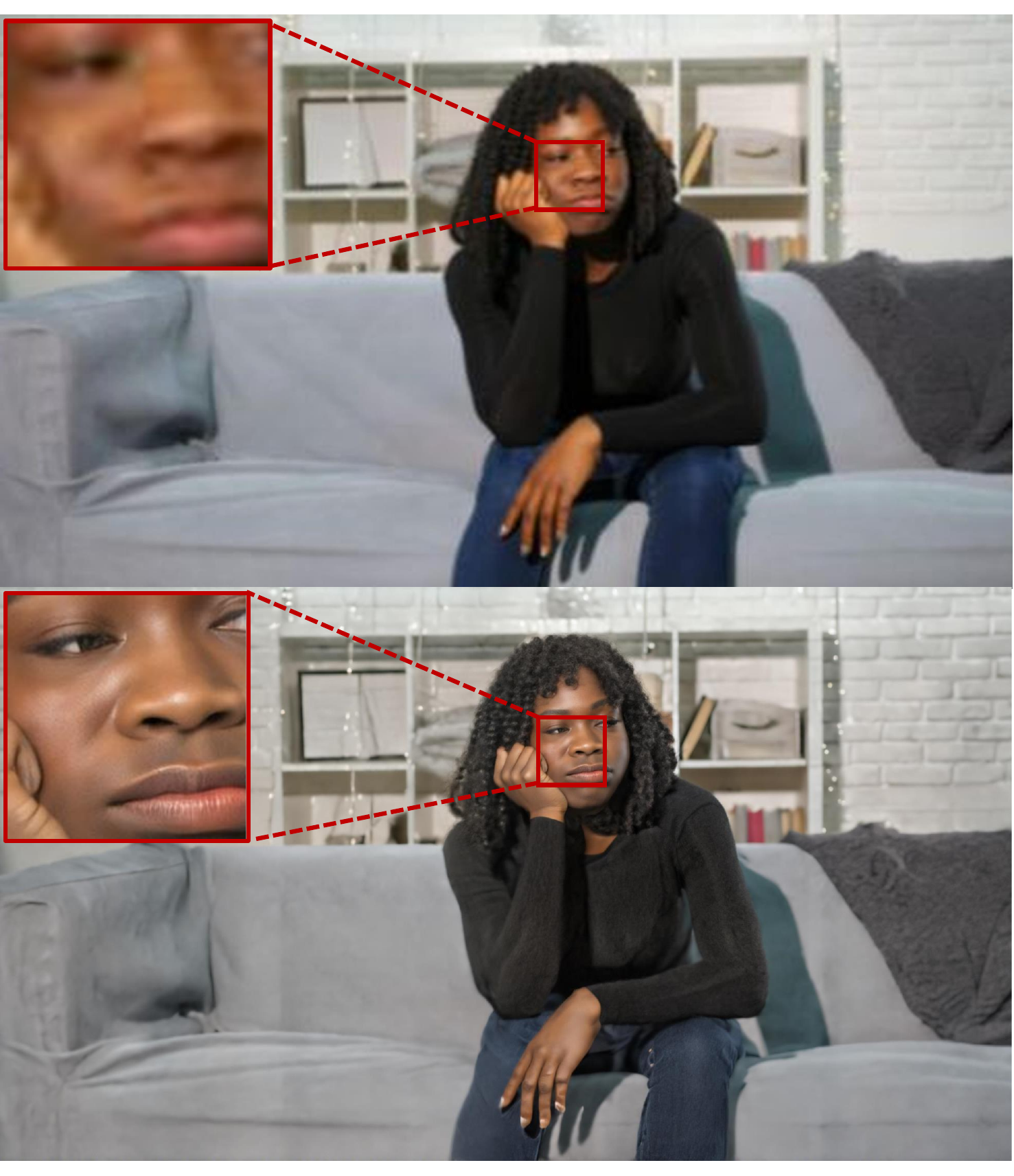}
    \caption{Example results on 4K image super-resolution (3648$\times$2048) .\textbf{Top:} Input low resolution image. \textbf{Bottom:} \model predicted high resolution image.}
    \label{fig:4k2}
\end{figure*}

\begin{figure*}
    \centering
    \includegraphics[width=0.9\linewidth]{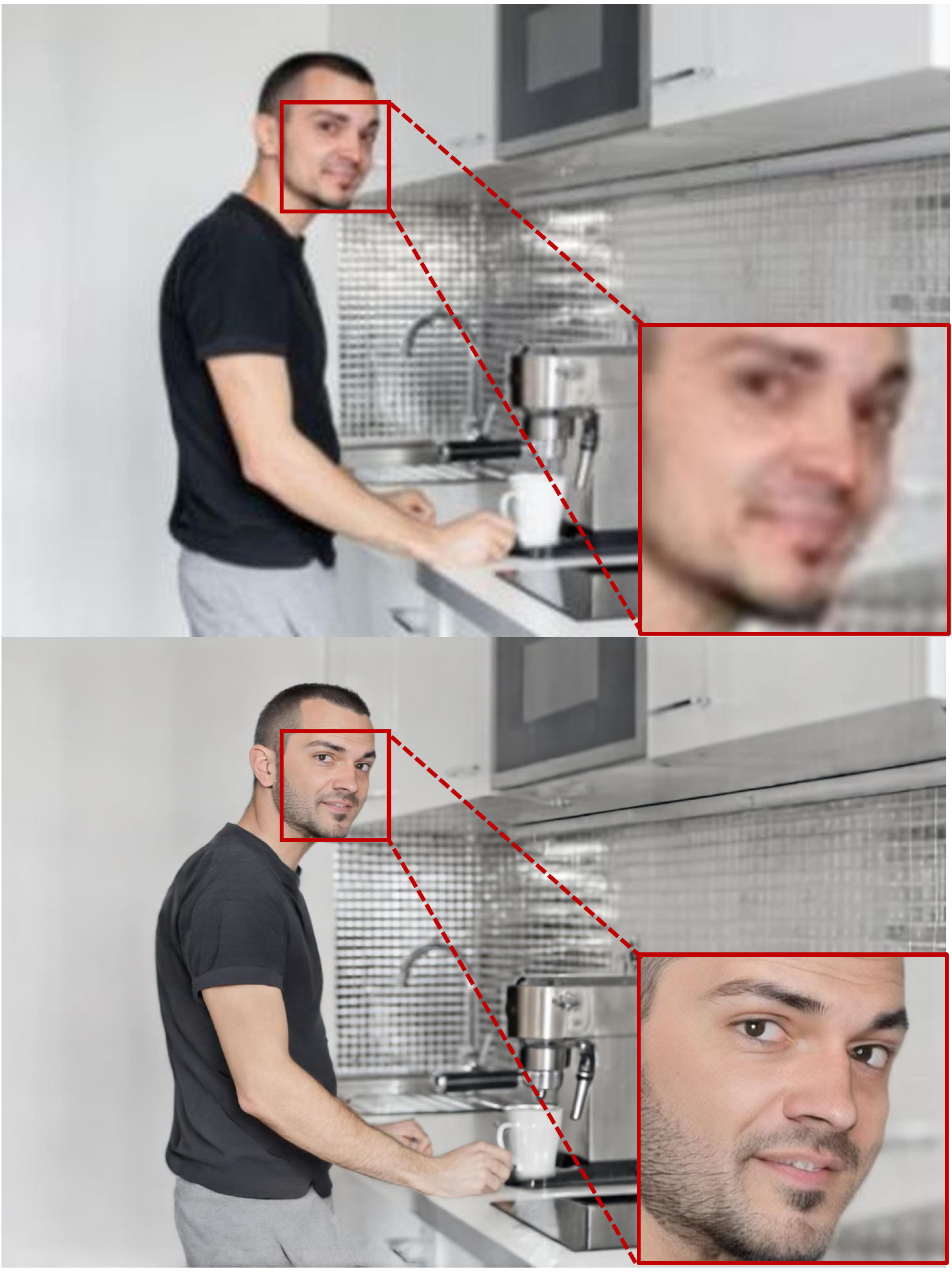}
    \caption{Example results on 4K image super-resolution (3072$\times$2048). \textbf{Top:} Input low resolution image. \textbf{Bottom:} \model predicted high resolution image}
    \label{fig:4k1}
\end{figure*}

\begin{figure*}
    \centering
    \includegraphics[width=\linewidth]{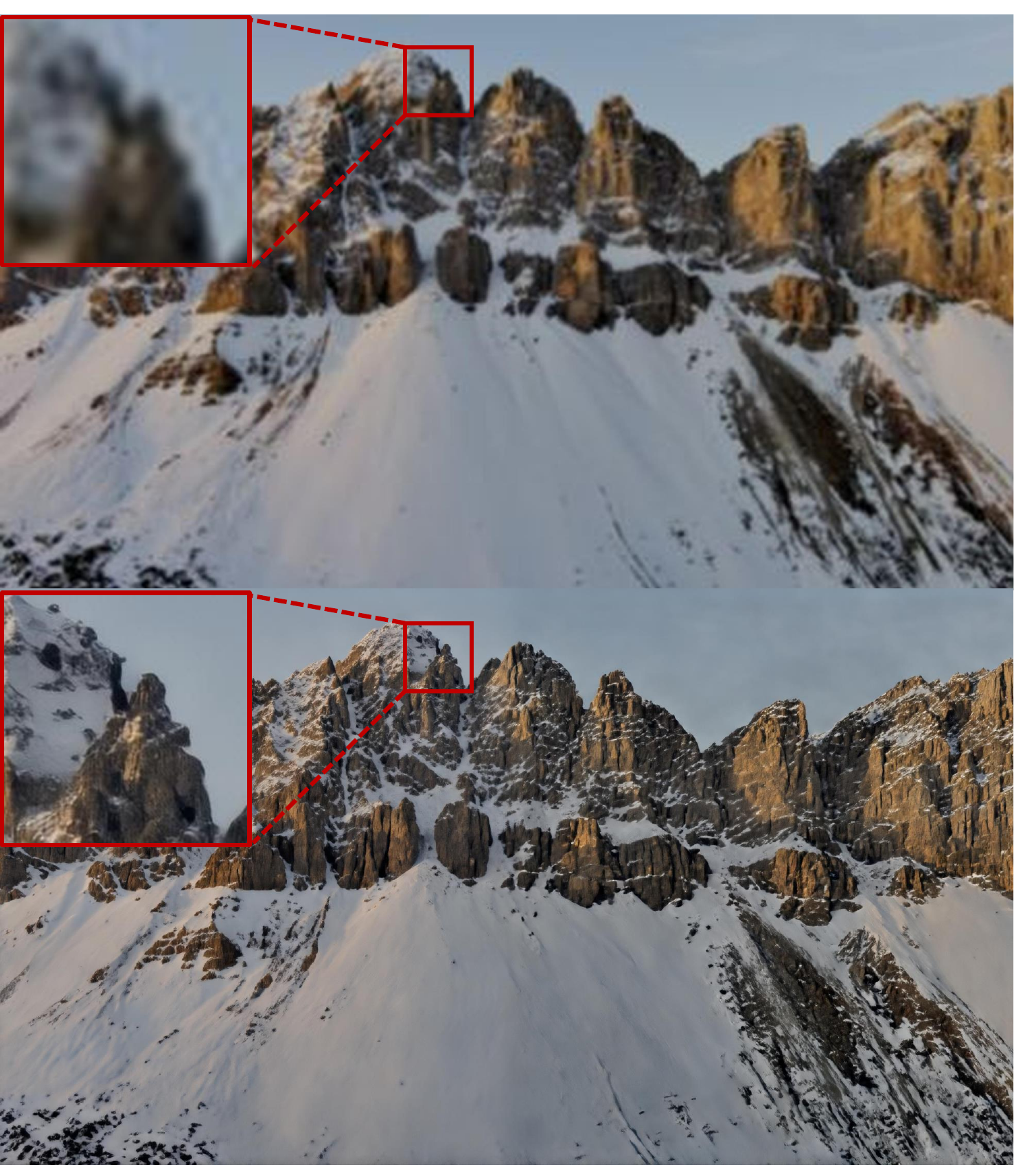}
    \caption{Example results on 4K image super-resolution (3648$\times$2048) .\textbf{Top:} Input low resolution image. \textbf{Bottom:} \model predicted high resolution image.}
    \label{fig:4k3}
\end{figure*}

\begin{figure*}
    \centering
    \includegraphics[width=\linewidth]{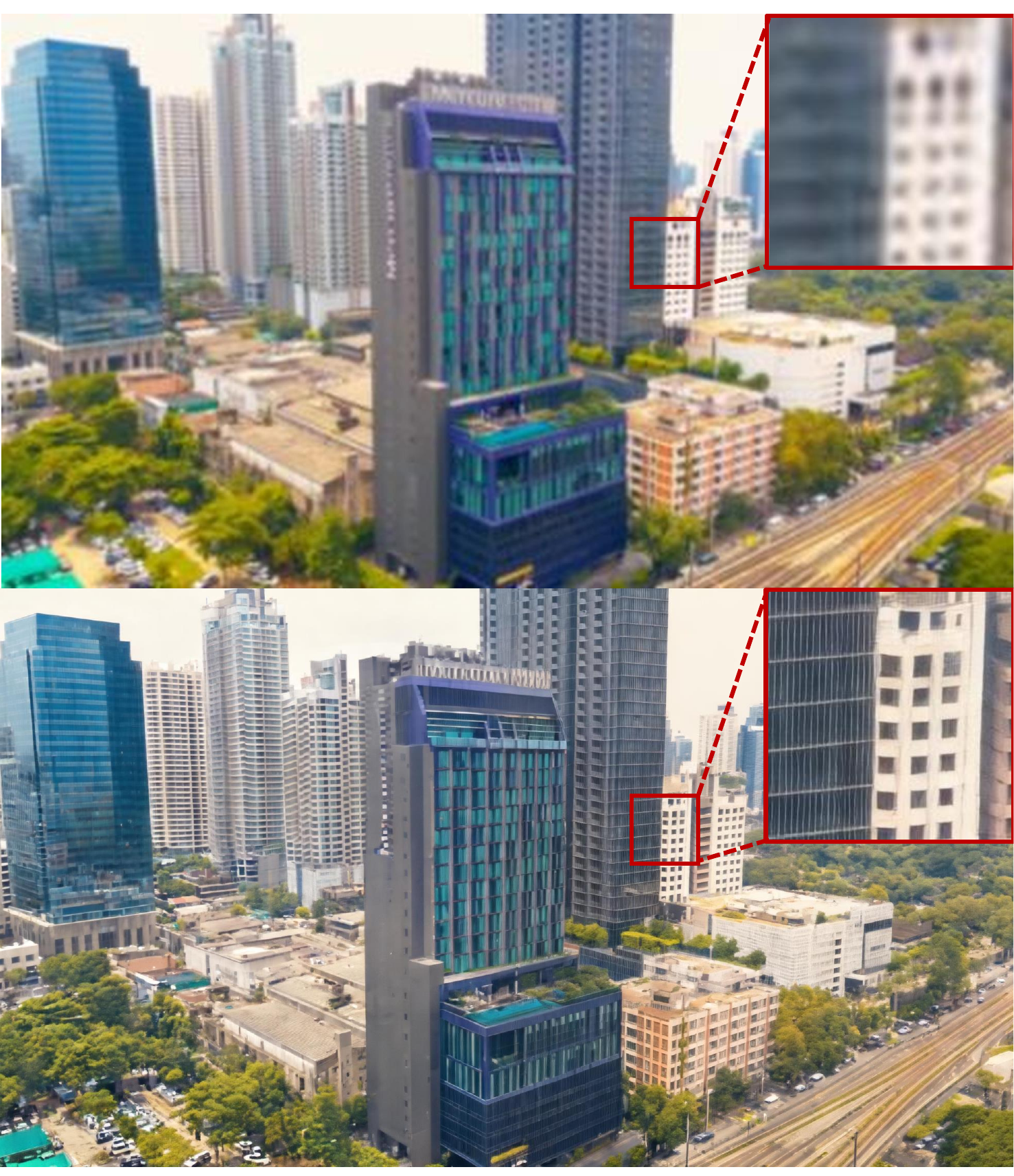}
    \caption{Example results on 4K image super-resolution (3648$\times$2048) .\textbf{Top:} Input low resolution image. \textbf{Bottom:} \model predicted high resolution image.}
    \label{fig:4k4}
\end{figure*}

\begin{figure*}
    \centering
    \includegraphics[width=0.9\linewidth]{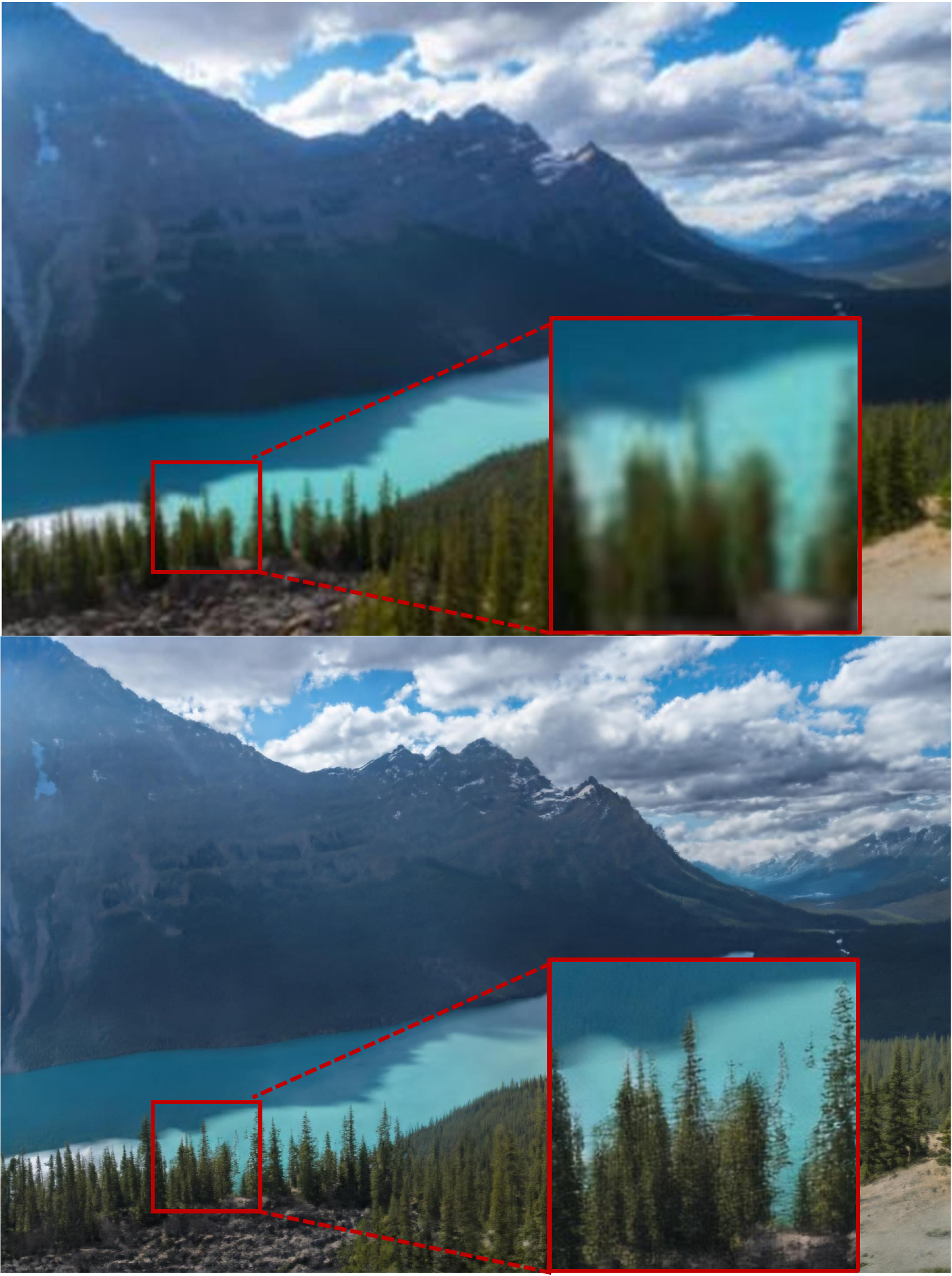}
    \caption{Example results on 4K image super-resolution (3072$\times$2048). \textbf{Top:} Input low resolution image. \textbf{Bottom:} \model predicted high resolution image}
    \label{fig:4k5}
\end{figure*}

In the second stage, we train on image SR, using model weights from the first stage as initialization. The training is conducted with a batch size of 256 for 30K iterations.
We show several results in Figures~\ref{fig:4k1}, \ref{fig:4k2}, \ref{fig:4k3}, \ref{fig:4k4}, and \ref{fig:4k5}. It can be observed that, in the SR task, our model performs especially well on portraits and also achieves remarkable results on landscapes and architecture. Overall, if a scene is handled well by the text-to-image generation model in the first stage, it is also effectively processed by the super-resolution model in the second stage.

\newpage

